\documentclass[letterpaper, 10 pt, conference]{ieeeconf}  % Comment this line out if you need a4paper%%%%%%%%%%%%%%%%%%%%%%%%%%%%%%%%%%%%%%%%%%%%%%%%%%%%%%%%%%%%%%%%%%%%%%%%%%%%%%%%

\IEEEoverridecommandlockouts                              % This command is only needed if 
\usepackage{xcolor}
\usepackage[hyphens]{url}
\usepackage{graphicx}%
\usepackage{multirow}%
\usepackage{amsmath,amssymb,amsfonts}%
\usepackage{mathrsfs}%
\usepackage{xcolor}%
\usepackage{textcomp}%
\usepackage{manyfoot}%
\usepackage{booktabs}%
\usepackage{algorithm}%
\usepackage{algorithmicx}%
\usepackage{algpseudocode}%
\usepackage{listings}%
\usepackage{subcaption}
\usepackage{footmisc}

\definecolor{darkred}{RGB}{180, 0, 0}
\definecolor{darkgreen}{RGB}{0, 180, 0}
\usepackage{pifont}% http://ctan.org/pkg/pifont
\usepackage{soul}
\newcommand{\cmark}{\color{darkgreen}\ding{51}}%
\newcommand{\xmark}{\color{darkred}\ding{55}}%

\newcommand{\unmarkedfootnote}[1]{%
    \begingroup
    \renewcommand{\thefootnote}{}%
    \footnotetext{#1}%
    \addtocounter{footnote}{-1}%  % Counter decrement to offset the increment by footnotetext
    \endgroup
}

\title{\LARGE \bf
Multi-person 3D pose estimation from unlabelled data
}

\author{Daniel Rodr\'iguez-Criado$^{1}$, Pilar Bachiller$^{2}$, George Vogiatzis$^{1}$ and Luis J. Manso$^{1}$% <-this % stops a space
\thanks{$^{1}$Daniel Rodriguez-Criado, George Vogiatzis and Luis J. Manso are with the College of
Engineering and Physical Sciences, Aston University, B4 7ET Birmingham,
UK
        {\tt\small \{190229717, g.vogiatzis, l.manso\}@aston.ac.uk}}%
\thanks{$^{2}$Pilar Bachiller is with the Robotics and Artificial Vision Laboratory,
University of Extremadura, Extremadura, Spain 
        {\tt\small pilarb@unex.es}}%
}

\begin{document}

\maketitle

\unmarkedfootnote{\textbf{This preprint has not undergone peer review. The Version of Record of this article is published in Machine Vision and Applications, and is available online at https://doi.org/10.1007/s00138-024-01530-6.}}
\thispagestyle{empty}
\pagestyle{empty}

%%%%%%%%%%%%%%%%%%%%%%%%%%%%%%%%%%%%%%%%%%%%%%%%%%%%%%%%%%%%%%%%%%%%%%%%%%%%%%%%

\begin{abstract}%   <- trailing '%' for backward compatibility of .sty file
Its numerous applications make multi-human 3D pose estimation a remarkably impactful area of research. Nevertheless, it presents several challenges, especially when approached using multiple views and regular RGB cameras as the only input.
First, each person must be uniquely identified in the different views.
Secondly, it must be robust to noise, partial occlusions, and views where a person may not be detected.
Thirdly, most pose estimation approaches rely on environment-specific annotated datasets that are frequently prohibitively expensive and/or require specialised hardware. Specifically, this is the first multi-camera, multi-person data-driven approach that does not require an annotated dataset.
In this work, we address these three challenges with the help of self-supervised learning.
In particular, we present a three-staged pipeline and a rigorous evaluation providing evidence that our approach performs faster than other state-of-the-art algorithms, with comparable accuracy, and most importantly, does not require annotated datasets.
The pipeline is composed of a 2D skeleton detection step, followed by a Graph Neural Network to estimate cross-view correspondences of the people in the scenario and a Multi-Layer Perceptron that transforms the 2D information into 3D pose estimations.
Our proposal comprises the last two steps, and it is compatible with any 2D skeleton detector as input.
These two models are trained in a self-supervised manner, thus avoiding the need for datasets annotated with 3D ground-truth poses.
\end{abstract}

\begin{keywords}
  3D multi-pose estimation, deep learning, graph neural networks, self-supervised learning
\end{keywords}

\section{Introduction}
% \textbf{Why developing a human-tracker system?}
Human detection and pose modelling have a plethora of applications, including video surveillance~\cite{Wang2013IntelligentMV}, assisted living~\cite{assistedLiving2011}, and autonomous vehicles~\cite{autonomousVehicles2021}.
In addition to any direct application, it is also the basis of trajectory prediction~\cite{shafiee2021introvert,sun20183dof}, interaction detection, and gesture recognition~\cite{aggarwal2014human}.
The number and relevance of applications make it extremely impactful.
Extensive research efforts have been made with different technologies such as LiDAR~\cite{lidar3D2017, lidar2D2011}, RGB cameras~\cite{tu2020voxelpose}, and RGBD cameras~\cite{Camplani2017,ZHANG201686}.

\par
There are multiple usability, cost and operational requirements that can be expected in a Human Pose Estimator (HPE).
First, most applications require support for more than one person.
Secondly, except for very few niche cases, HPEs must work with occluded body parts.
Most applications will also benefit from limiting sensors to RGB cameras, avoiding RGBD or other more expensive hardware.
An ideal HPE would also exploit the available context to provide 3-dimensional data for all keypoints, even if not all of them are visible.
Another desirable feature for any learning-based HPE would be not to require a labelled dataset to be implemented in a new space, as they are expensive to compile.
\par

% \textbf{Why machine learning?}
RGB-based multi-human and multi-view 3D pose estimation is usually done in three steps: a) detect humans and estimate their 2D poses on the images using, for example, a Convolutional Neural Network (CNN); b) search for correspondences in the different views of the people detected in the previous step; and c) estimate 3D poses for each person based on the image coordinates of their keypoints for the different views.
This work builds on top of publicly available pose detectors (\textit{e.g.}, \cite{matterportmaskrcnn2017, openpose2019}) for the first step of the pipeline and presents a novel solution for the second and third steps.
It is important to note that the system developed in this research can be integrated with any third-party 2D detector.
\par

Regarding the second step, which consists on associating the 2D poses that correspond to the same person in the different images, the literature addresses the problem using both appearance and geometry cues.
Examples of this are the use of epipolar geometry to assign a cost to each pose detected~\cite{9025555} or the embedding of appearance features using a pre-trained model to provide affinity scores between bounding boxes~\cite{dong2021fast}.
Due to the desired multi-person support and the irrelevance of the order in which people are detected, we chose to exploit Graph Neural Networks (GNNs) to match people's views, as they are order-invariant and can manage a variable number of input nodes.
\par

Traditionally, the final step, 3D pose estimation, has been done using triangulation or pictorial structure models.
The main limitation of these classic approaches stems from the inability to predict the occluded parts, as these methods are not capable of estimating positions for keypoints that are occluded in many or all views.
To overcome these limitations, learning-based solutions have emerged.
It can be argued that an artificial neural network can learn to \textit{hallucinate} the occluded parts of the body even if they are not visible. 
This is based on the intuition that the network should be able to exploit contextual information from the rest of the keypoints and the existing views, if any.
For instance, a network could learn to implicitly internalise the proportions of the human body and its bilateral symmetry.
Therefore, if the keypoint for the left elbow cannot be seen from any camera, knowing the position of the wrist and the average proportions of a human forearm (or the length of the opposite forearm), the network could estimate the position of the elbow.
Embedding these complex but helpful biases efficiently would be very challenging in non-data-driven approaches.
\par

% \textbf{Why self-supervised learning?}
A significant limitation of most current data-driven solutions, and more importantly, all of those that provide multi-camera support for multi-person pose estimation, is the necessity of annotating the datasets to train the models in a supervised fashion.
It is worth noting that multi-camera datasets are specific to the relative positions of the cameras, making the datasets scenario-specific.
As a consequence, to use the corresponding approaches, an annotated dataset has to be compiled for every scenario, which is time-consuming and requires expensive tracking systems.
\par
In addition, while it is feasible to utilise 3D data from an in-studio dataset to establish 2D-3D relationships for different camera configurations via 3D projections, a model trained on such a dataset would exhibit significant sensitivity to variations in the 2D detected keypoints.
This sensitivity arises from training the model using ideal 2D coordinates, in contrast to the potentially noisy 2D coordinates used during inference.
Even assuming that the 2D keypoint detection is noise-free, training using ground truth data will likely fail at inference time due to ground truth keypoint projection coordinates not matching the coordinates of the skeleton keypoints considered by the 2D detectors.
Another limitation of this dataset generation approach is related to the variability of the dataset.
Depending on the application, it could be necessary to have data on individuals with diverse complexion, heights, and even ages, which might not be readily available in an in-studio dataset.
The process of gathering such diverse data would essentially take us back to the initial challenge: obtaining an annotated dataset.
\par
To deal with these problems and avoid the need for annotated datasets, we propose a self-supervised learning-based solution, with two main contributions:
\begin{itemize}
\item An elegant solution for matching different 2D poses from several cameras using a GNN that allows having a variable number of people in the scenario.
\item A model that infers the 3D keypoints of the detected humans using self-supervised learning by minimising the difference between the 2D detected keypoints' coordinates and those of the estimated poses' re-projections. 
\end{itemize}

\par
The following sections cover various aspects of our work.
Section \ref{sec:related_work} reviews relevant 2D human pose detectors and presents the current state of 3D pose estimation.
The proposed method is described in detail in Section~\ref{sec:method}.
Section \ref{sec:results} presents experimental results, including a performance comparison with other state-of-the-art methods, using two distinct datasets. 
Additionally, we will show how the system can be applied to mobile robots without retraining for different scenarios as long as only on-board cameras are used.
Finally, Section~\ref{sec:conclusions} summarises the main conclusions.

%%%%%%%%%%%%%%%%%%%%%%%%%%%%%%%%%%%%%%%%%%%%%%%%%%%%%%%
%%%%         R E L A T E D     W O R K            %%%%%
%%%%%%%%%%%%%%%%%%%%%%%%%%%%%%%%%%%%%%%%%%%%%%%%%%%%%%%
\section{Related work}
\label{sec:related_work}
This section provides an overview of the leading literature on 3D Human Pose Estimation. 
We start with a brief discussion of popular 2D detectors, as they are leveraged in various 3D estimation models -including ours. 
We omit works that utilise RGB-D sensors, since our work focuses on RGB cameras, offering the advantage of significantly reduced equipment costs.
\par
\textbf{2D human pose estimators} yield image coordinates of human anatomical keypoints in an image for every detected person. 
Recent advancements in deep learning have led to a significant improvement in the performance and accuracy of these models, surpassing the previous approaches that relied on probabilistic and hand-crafted features~\cite{Chen2022}.
Most of these learning-based models~\cite{matterportmaskrcnn2017, Kreiss2019, Jain2014, Tompson2014} rely on Convolutional Neural Networks.
% (LUIS CHANGE)
There is a vast number of 2D pose estimators, with OpenPose~\cite{openpose2019} being one of the most popular.
It leverages part affinity fields for human parts association using a bottom-up approach.
A similar approach is followed by OpenPifPaf~\cite{kreiss2021openpifpaf} and trt-pose\footnote{\url{https://github.com/NVIDIA-AI-IOT/trt_pose}}.
Another widely known 2D pose detector is HRNet~\cite{Sun2019}, which can maintain high-resolution representations through the detection process, claiming higher accuracy and spatial precision.
One of the most popular datasets used for training and evaluating these 2D models is COCO~\cite{Lin2014}, containing more than 100.000 annotated images. 
\par
In relation to the \textbf{3D pose estimation} problem, fuelled by the outstanding advances in 2D estimations, many works have tried to utilise these models for estimating 3D poses from the 2D points~\cite{Wang2021}.
Many of them retrieve 3D human poses from monocular views~\cite{li20143d,rogez2019lcr,XNect_SIGGRAPH2020,Pavlakos2017,Moreno-Noguer2017}, although they suffer from the unavoidable fact that monocular depth estimation is an ill-posed problem, as multiple potential 3D poses are possible given a single 2D projection.
\textbf{Multi-camera} systems can reduce this ambiguity significantly, and increase their robustness against occlusions and noise. 
However, multi-view Human Pose Estimation with \textbf{multiple people} introduces the challenge of matching each person's set of keypoints among the images of the different cameras.
Previous works have addressed this problem with algorithms based on appearance and geometric information~\cite{dong2021fast, Belagiannis2016}.
\cite{dong2021fast} create affinity matrices based on the appearance between two views and use them as input to their model to infer the correspondence matrix. 
\par

Once the cross-view correspondences are solved, there are several techniques to merge the information from the different views to extract the 3D pose.
Most classical approaches rely on epipolar geometry by triangulating the 3D points from the 2D points~\cite{amin2013multi, Belagiannis2016, Kocabas2019}.
The pictorial structure paradigm was extended to 3D to deal with multi-human pose detection in~\cite{6909612}. 
However, the model does not detect full skeletons in case of occlusions and they assume to know the number of people in the scene, which is not a realistic assumption~\cite{tu2020voxelpose}.
Other works tackle the problem with prediction models based on \textbf{deep learning} and CNNs~\cite{tu2020voxelpose, ye2022faster, rodriguez2020multi}.
For example, VoxelPose~\cite{tu2020voxelpose} discretises the 3D space in small cubes called voxels.
Using this representation, the 2D heatmaps detected from all the views are projected into a common 3D space and two 3D convolutional models are applied.
The first model yields detection proposals for each person and the second estimates the positions of the keypoints for each proposal.
This method avoids establishing cross-view correspondence based on poor-quality 2D poses.
\textit{Ye et al.}~\cite{ye2022faster} present an accelerated version of VoxelPose which avoids the use of 3D convolutions, although the results are marginally worse. 
Firstly, they re-project the aggregated feature volume, which is acquired in the same way as in~\cite{tu2020voxelpose}, to the ground plane ($xy$) by implementing max-pooling along the $z$-axis.
Next, they employ a 2D-CNN network over the $xy$-plane to locate individuals and generate a 1D feature vector in the $z$-axis for each detection.
Finally, they apply a 1D-CNN to that vector to get the final 3D pose estimation.
These modifications enable their model to achieve results approximately $10$ times faster without sacrificing precision.
Another interesting work is presented by~\cite{lin2021multi}.
Their approach utilises a plane sweep stereo technique to simultaneously address the challenges of multiple-view fusion and 3D pose estimation.
All these models use supervised learning, thus they require datasets with precise annotations.
% It is worth noting that all these models are camera-configuration specific and require scenario-specific datasets with precise annotations.
% Overcoming this second issue is the \textbf{main aim} of this work.
% (LUIS CHANGE)
The number of these datasets is scarce mainly due to the costly equipment required as well as the need for a controlled environment to record the data.
Some examples are the Human3.6M dataset~\cite{h36m_pami}, with more than 10 thousand annotations from 1 thousand images, and the CMU Panoptic dataset~\cite{panoptic}, with 5.5 hours of video from different angles and 1.5 million of 3D annotated skeletons.
\par

Besides CNN-based models, multiple works in the literature address the problem with the use of GNNs.
For example, works such as~\cite{xu2021invariant, hu2021conditional} obtain promising results from monocular views using GNNs.
\cite{wu2021graph} propose a solution for multi-view and multi-person 3D estimation using GNNs with supervised learning for both, cross-view correspondence and final 3D pose estimation.
They construct the graphs by transforming each detected keypoint into a graph node and use the natural connections in the body to generate the graph edges.
Then the GNN applies a regression in the node features to obtain the 3D coordinates of the body joints. 
The main limitation is that the training of these networks requires datasets with accurate 3D ground truth annotations.

\begin{table*}
\centering
\caption{Qualitative comparison with literature}
\label{tbl:qualityComparison}
\resizebox{0.9\linewidth}{!}{%
\begin{tabular}{c|ccc}
\textbf{Reference}                                       & \textbf{Multi-camera} & \textbf{Multi-person} & \textbf{Self-supervised} \\
\hline
Tu et al. (2020) \cite{tu2020voxelpose}          &\cmark          &\cmark          & \xmark              \\
Ye et al. (2022) \cite{ye2022faster}             &\cmark          &\cmark          & \xmark              \\
Wu et al. (2021) \cite{wu2021graph}              &\cmark          &\cmark          & \xmark              \\
Lin and Li (2021) \cite{lin2021multi}            &\cmark          &\cmark          & \xmark              \\
\hline
Biswas et al. (2019) \cite{biswas2019lifting}    & \xmark           & \xmark           &\cmark             \\
Kundu et al. (2020) \cite{kundu2020self}         & \xmark           & \xmark           &\cmark             \\
Srivastav et al. (2020) \cite{srivastav2020self} & \xmark           &\cmark          &\cmark             \\
\hline
Bouazizi et al. (2021) \cite{Bouazizi2021}    & \cmark           & \xmark           &\cmark             \\
Bartol et al. (2022) \cite{Bartol2022}         & \cmark           & \xmark           &\cmark             \\
Bala et al. (2023) \cite{Bala2023} & \cmark           &\xmark          &\cmark             \\
Gong et al. (2023) \cite{Gong2023} & \cmark           &\xmark          &\cmark             \\
\hline
Ours                                                       &\cmark          &\cmark          &\cmark             \\
\hline
\end{tabular}
}
\end{table*}

Due to the remarkable benefits of avoiding 3D annotations (\textit{i.e.}, datasets become much more affordable in terms of cost and effort, which facilitates collecting larger datasets), self or semi-supervised learning methods have been proposed in many works~\cite{Drover2019, Pavlakos2017, Rhodin2018, Kocabas2019, biswas2019lifting, kundu2020self, srivastav2020self, Bouazizi2021, Bartol2022, Gong2023}.
These methods primarily focus on single-view 3D pose estimation or are constrained to single individuals. To the best of our knowledge, our proposal stands as the first multi-camera 3D Human Pose Estimation method that supports multiple individuals without requiring ground truth data. A qualitative comparison of recent works is presented in Table~\ref{tbl:qualityComparison}, showing that our method is the only one that meets these three criteria.
\par

%%%%%%%%%%%%%%%%%%%%%%%%%%%%%%%%%%%%%%%%%%%%%%%%%%%%%%%
%%%%              M  E  T  H  O  D                %%%%%
%%%%%%%%%%%%%%%%%%%%%%%%%%%%%%%%%%%%%%%%%%%%%%%%%%%%%%%
\section{Method}
\label{sec:method}
\begin{figure*}[!ht]
\centering
\includegraphics[width=0.8\textwidth]{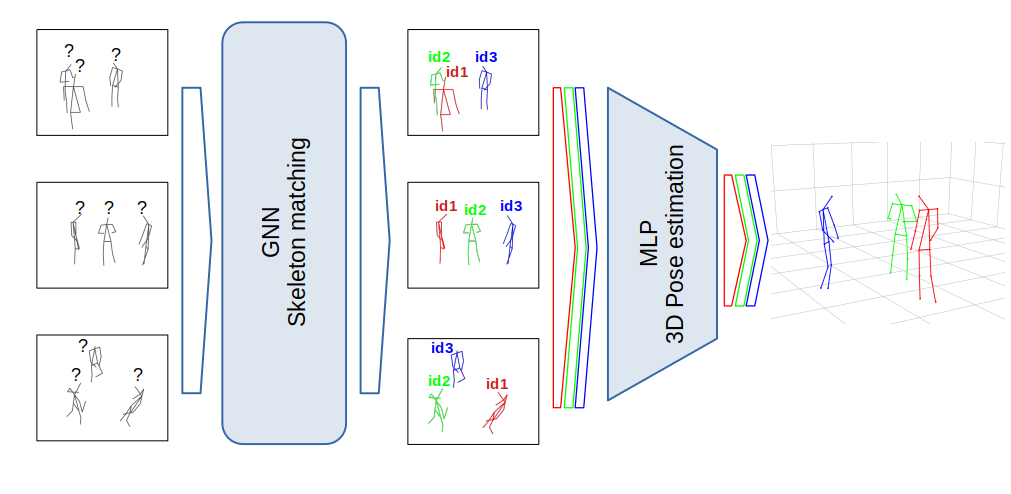}
\caption{Two last stages of the pipeline of the proposed system. The correspondences between the input skeletons in the different views are estimated by the GNN. This information is leveraged by the MLP to provide the final 3D poses.}
\label{fig:pipeline}
\end{figure*}

The proposed system consists of a three-staged pipeline: a) a skeleton detector, b) a multi-view skeleton matching Graph Neural Network, and c) a pose estimation Multi-Layer Perceptron (MLP).
Given that there are very efficient solutions for the first stage of the pipeline, no new alternative is proposed in this work.
In fact, our system is independent of the skeleton detector used.
The multi-view skeleton matching and the pose estimation network are our two main contributions.
Figure~\ref{fig:pipeline} shows how these two stages of the pipeline work at test time, which take as input a set of detected skeletons per view that can be obtained using any skeleton detector.
The code is available at \url{https://github.com/gnns4hri/3D_multi_pose_estimator}.

\subsection{System calibration}
\label{img_processing}
Our proposal is not limited to a number or configuration of cameras but requires the camera configuration to be the same during the collection of the dataset and the final inference for pose estimation.
The only exception to this is that our system allows the set of cameras used at inference time ($\mathbb{C}_i$) to be a subset of the cameras used during training ($\mathbb{C}_t$).
In that case, only the cameras in $\mathbb{C}_i$ need to maintain the configuration they had at training time during inference time.
If desired, the rest of the cameras in $\mathbb{C}_t$ can be removed from the system once it is trained.
This is particularly helpful for two reasons: a)~it allows to better \textit{estimate} the 3D positions of keypoints that are occluded or not detected at inference time by the cameras in $\mathbb{C}_i$; and b)~it improves inference time accuracy when $\mathbb{C}_t$ can be higher than $\mathbb{C}_i$ (\textit{e.g.}, in mobile robots that need to use a small set of cameras at inference time, but can use additional cameras when compiling the training dataset).
% This is particularly helpful for two reasons: a)~it improves inference time accuracy when $\mathbb{C}_t$ can be higher than $\mathbb{C}_i$ (\textit{e.g.}, in mobile robots that need to use a small set of cameras at inference time, but can use additional cameras when compiling the training dataset); and b)~it allows to better \textit{imagine} the 3D positions of keypoints that are not detected at inference time by the cameras in $\mathbb{C}_i$.
\par
The first step to set up the system is to calibrate the intrinsic parameters of all the cameras available, as well as their extrinsic parameters with respect to the desired global frame of reference. Using these parameters, the projection matrices of all the cameras ($T^c  \; \forall c \in \mathbb{C}_t$) are created.
These matrices are used during the training and inference phases of the two proposed neural networks, as described in sections~\ref{sec:skeleton_matching} and~\ref{MLP_pose}.
\par

\subsection{Skeleton detection}
For training purposes, once the system has been calibrated, a dataset specific to the camera configuration used at training time ($\mathbb{C}_t$) needs to be collected.
The training works on the assumption that the dataset has been generated with a single person in the environment at a time.
This does not apply to inference time, where the number of people is not theoretically bounded.
The detected skeletons are represented as a list of keypoints, defined by their 2D image coordinates along an identifier for each skeleton keypoint and a certainty value for the detection.
Datasets contain sequences of samples, each consisting of a list of the detected skeletons per camera.
\par
% (LUIS CHANGED)
As aforementioned, our proposal can be used with any skeleton detector, regardless of the number of keypoints they provide for each skeleton.
The number of keypoints only determines the size of the input features of each network, so it is a mere configuration parameter.

\subsection{Skeleton matching}
\label{sec:skeleton_matching}
Once the skeletons are detected in the different views, the skeletons belonging to the same person are matched.
% (LUIS CHANGED)
Since the matching is expected to be order invariant and the number of people is unknown at inference time, we train a GNN model to estimate the correspondence between all the views.
This is because GNNs are order-invariant and allow for a variable number of input nodes.
\par

One of the main limitations of learning-based approaches using supervised learning is the generation of the dataset, which needs to be annotated with ground truth.
For this particular problem, if the dataset contained more than one skeleton at a time, it would be necessary to manually annotate their cross-view correspondence.
To avoid this arduous process, our raw training dataset contains a set of sequences of single individuals moving around the environment -one at a time.
These sequences can then be combined into a single processed dataset that contains ground truth labels built by aggregating the data of multiple individuals.
We describe this process below.
\par 

Our GNN model receives as input an undirected graph $G=(V,E)$, where $V$ is the set of nodes, and $E$ is the set of edges. 
The set $V$ is composed of two different types of nodes that we term \textit{detection} nodes and \textit{match} nodes.
These two types of nodes are the elements of $V_{d}$ and $V_{m}$ respectively, such that $V=V_{d}\cup V_{m}$.
Each \textit{detection} node represents a 2D skeleton detected in one of the views used at inference time (i.e. a view $c \in \mathbb{C}_i$), while a \textit{match} node represents a possible match between two different detections.
It is worth noting that only the cameras in $\mathbb{C}_i$ are used because we are only interested in these cameras at inference time and including the rest of the cameras in $\mathbb{C}_t$ would not add any valuable information to this end.
For each pair of detection nodes $v_{i}, v_{j} \in V_{d}$ such that $v_{i}$ and $v_{j}$ belong to different views, there is a corresponding match node $v_{k} \in V_{m}$. 
The edges in $E$ connect the match nodes to their corresponding detection nodes. Thus, for each match node $v_{k} \in V_{m}$ linking two detection nodes $v_{i}, v_{j} \in V_{d}$, there are two edges $(v_{k}, v_{i})$ and $(v_{k}, v_{j})$.
Therefore, the input graph $G$ can be represented as follows:
\begin{equation}
G = (V_{d}\cup V_{m}, (v_{k}, v_{i}) \cup (v_{k}, v_{j}))
\end{equation}
\noindent where $v_{i}, v_{j} \in V_{d}$ and $v_{k} \in V_{m}$ correspond to the match node between detection nodes $v_{i}$ and $v_{j}$.
\par

Each node (\textit{detection} or \textit{match}) is represented with a feature vector ($x$) with $N_k \times N_c \times 10 + 2$ elements, where $N_k$ and $N_c$ are the number of keypoints and cameras, respectively. 
Two of these elements denote a binary 1-hot encoding indicating if the node is a \textit{detection} or a \textit{match}. In the case of a \textit{match} node, all other dimensions are fixed to zero. In the case of a \textit{detection} node there is a 10-dimensional tuple for each camera-keypoint combination, each of which consists of:
\begin{itemize}
    \item a flag indicating if the keypoint has been detected,
    \item the pixel coordinates if the keypoint is visible (2 zeros otherwise),
    \item a value within the range [0, 1] indicating the certainty of the detection of the keypoint (zero if the keypoint is not visible),
    \item six elements encoding the 3D line passing through the origin of the camera and the keypoint (image plane coordinates) in the global frame of reference (specified as a 3D point and a 3D direction vector).
\end{itemize}
\par

Given the input graph $G=(V,E)$ and the feature vectors of the set of nodes ($x_i\in\mathbb{R}^d \text{, for }v_i \in V$), the GNN is trained to produce an output graph $G'=(V,E)$ with the same structure as $G$, but different feature vectors for each node ($y_i\in\mathbb{R}^{d'}$).
In particular, it is trained to predict whether each \textit{match} node $v_{k} \in V_m$ corresponds to a true match.
Thus, we formulate the matching as a binary classification task, where the target labels are $\{0,1\}$ for non-matches and matches respectively.
To ensure that each output of the GNN is within the range of $[0, 1]$, we use the Sigmoid activation function in the output layer. 
Consequently, both the binary cross-entropy (BCE) loss and the mean squared error (MSE) loss are suitable for computing the loss during the training of the GNN.
In our experimental results, the GNN trained with MSE loss yields slightly better performance than the GNN trained with BCE loss. 
Therefore, we define the GNN loss in terms of MSE loss:
\begin{equation}
    \mathcal{L}_{SM} = \frac{1}{\lvert V_m \rvert}\sum_{v_k \in V_m}(y_{k}-\hat{y}_{k})^2
\end{equation}
\noindent being $y_k \in \{0,1\}$ the target label for node $v_{k} \in V_m$, and $\hat{y}_k\in[0,1]$ the predicted probability that node $v_{k}$ corresponds to a match.
\par

\begin{figure*}[ht!]
\centering
  \begin{subfigure}{0.325\textwidth}
    \centering
    \includegraphics[width=1\textwidth]{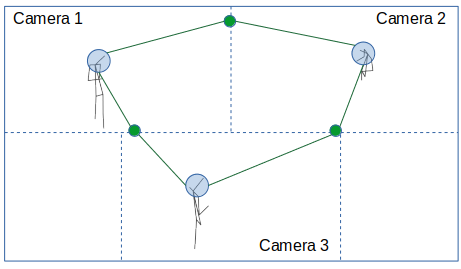}
    \caption{Graph for subject 1.}
    \label{fig:graph_matching_P1}
  \end{subfigure}
  \begin{subfigure}{0.325\textwidth}
    \centering
    \includegraphics[width=1\textwidth]{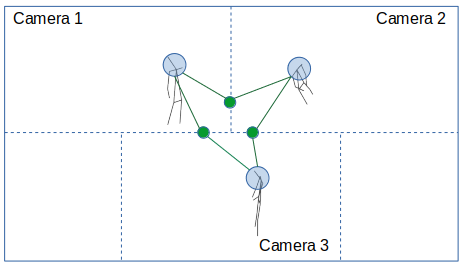}
    \caption{Graph for subject 2.}
    \label{fig:graph_matching_P2}
  \end{subfigure}
  \begin{subfigure}{0.325\textwidth}
    \centering
    \includegraphics[width=1\textwidth]{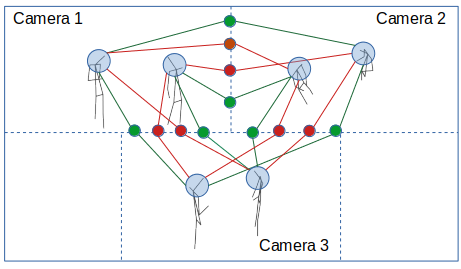}
    \caption{Graph for subject 1 and 2.}
    \label{fig:graph_matching_P1_and_P2}
  \end{subfigure}
  \caption{Generation of a sample of the dataset. Graphs of individual persons are generated first assigning a score of $1$ to the \textit{match} nodes connecting the views (green nodes). Then a final graph is generated from the individual ones adding \textit{match} nodes with a score of $0$ (red nodes).}
  \label{fig:graph_matching_dataset}
\end{figure*}

As mentioned previously, to avoid manual labelling, we use footage of single individuals walking and moving through our system.
Since each frame contains only one person, we can readily identify matching 2D detections. 
Using this data, we generate separate graphs for each person, where all \textit{match} nodes are assigned a maximum score value (see figures \ref{fig:graph_matching_P1} and \ref{fig:graph_matching_P2}).
We then combine the graphs of individual persons by adding \textit{match} nodes with a score of 0 connecting pairs of detections of different persons, as depicted in figure \ref{fig:graph_matching_P1_and_P2}.
By following this procedure, we generate the target label $y_k$ of each \textit{match} node $v_k \in V_m$ of the graphs composing the training set, allowing us to train the GNN in a pseudo-supervised manner.
The number of individual graphs to be combined is randomly determined for each sample in the dataset, with a minimum of one and a maximum equal to the total number of sequences used to generate the dataset.

\subsection{3D Pose estimation} \label{MLP_pose}
Having identified the different views of each person, an MLP is used to estimate the 3D coordinates of the keypoints for each of them.
The input features of the model are the concatenation of $14$ features per keypoint and camera. 
Therefore, if the skeleton detector detects up to $25$ keypoints and the system uses $3$ cameras at inference time ($\lvert \mathbb{C}_i\rvert =3$), the dimension of the input feature vector would be $14 \times 3 \times 25$, $1050$ dimensions in total. 
The $14$ features per keypoint correspond to the 10 features described in the previous section for skeleton matching plus four additional features related to an initial estimation of the 3D. 
Specifically, if a keypoint of a person is detected by 2 or more cameras, its 3D coordinates are reconstructed by triangulation for every pair of cameras and an initial estimation is computed as the centroid of the obtained 3D points.
This estimation is included as input using three of the four new features.
The last feature is used to indicate the availability of the estimated 3D.
It is set to $1$ if there is more than one view of the keypoint and to $0$ otherwise.
\par

Using the aforementioned information per keypoint and camera, the network estimates the 3D positions of the keypoints in the global frame of reference, yielding $x$, $y$ and $z$ for each of them. 
Thus, assuming that the network predicts the position of $25$ different keypoints, the output vector dimension is $3 \times 25 = 75$.
% (LUIS CHANGED)
% Bear in mind that an MLP query is performed per person, but the queries can be batched.
\par

The training process, as explained in the introduction, follows a self-supervised learning approach, which represents the main advantage of this approach.
This way, there is no need to use a ground truth to compare the output, since the loss function only uses the data from the skeleton detectors.
However, calculating this loss is not trivial, since the network infers 3D poses from 2D positions.
Our approach to solving this problem is to project the 3D coordinates of the keypoints predicted by the network into each camera used for training ($\mathbb{C}_t$).
The transformation between global and image coordinates is done by using the projection matrices ($T^c \; \forall c \in \mathbb{C}_t$) obtained during the calibration process.
Using the projected coordinates and the coordinates yielded by the skeleton detector, a measurement of the estimation error of the network is obtained.
This error defines the loss function that the network is trained to minimise.
More formally, assuming that the output of the network $o$ is represented as a vector of 3D positions corresponding to the estimation of the subject's keypoints coordinates:
\begin{equation}
o := \left ( o_{0}, o_{1}, ...,  o_{N_k-1} \right )
\label{mlp_output}
\end{equation}
\noindent with $N_k$ the number of keypoints, a vector $p^c$ of image projected positions ($p^c_{i}$) can be obtained for each camera as follows\footnote{The conversions between homogeneous and standard coordinates are omitted for simplification}:

\begin{equation}
p^c_{i} = T^{c} \cdot o_{i} \quad \forall i \in [0, N_k)
\label{camera_coor}
\end{equation}

Using $p^{c}$ and the set of detected keypoints ($S^c=\{s^c_{k}\}$) for each camera $c$, the projection error $e$ is computed as

\begin{equation}
e = \sum_{c \in \mathbb{C}_t}\sum_{s^c_{k} \in S^c} d(p^c_{k}, s^c_{k})
\label{projection_error}
\end{equation}

\noindent being $d(\cdot)$ the Manhattan distance between the projected and detected points.

Applying equation \ref{projection_error} to each sample of the dataset $D$, the final loss is calculated using the mean squared error:

\begin{equation}
\mathcal{L}_{3D} = \frac{1}{\lvert D \rvert}\sum_{d \in D} e_d^2
\label{mlp_loss}
\end{equation}

\noindent being $e_d$ the result of equation \ref{projection_error} for the sample $d$.

\begin{figure*}[!ht]
\centering
\includegraphics[width=0.85\textwidth]{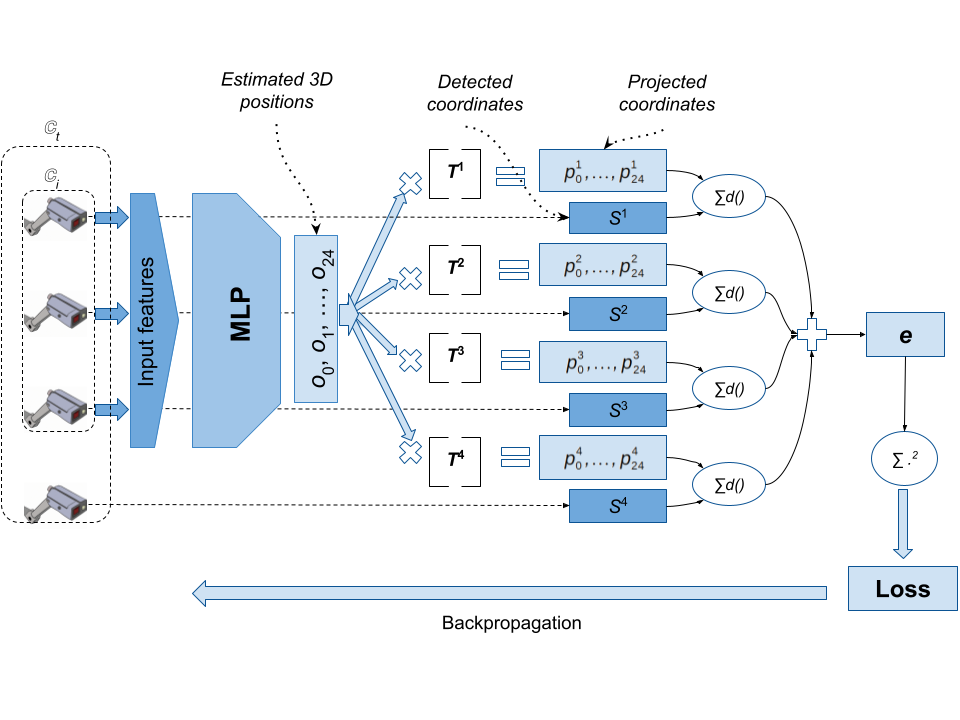}
\caption{Representation of how the 3D pose estimation network training loss is computed in a setup with $4$ cameras in $\mathbb{C}_t$ and $3$ cameras in $\mathbb{C}_i$.}
\label{fig:mlp_diagram}
\end{figure*}

Figure \ref{fig:mlp_diagram} depicts the process to compute the self-supervised loss, assuming $25$ keypoints, $4$ cameras in $\mathbb{C}_t$, and $3$ cameras in $\mathbb{C}_i$.
It can be observed that the loss computation utilises detections from all cameras in $\mathbb{C}_t$ but only the cameras in $\mathbb{C}_i$ are used for generating the network's input.
Consequently, in the aforementioned example, the model receives detections from only three of the four cameras at inference time.
However, even though the fourth camera would not be used at inference time, our HPE would still exploit what was learned from it at training time.

 \subsection{Data augmentation}
Data augmentation is applied to extend the data used for training, to increase the variety of situations, and to increase the robustness against partial views.
Specifically, for each original sample of the dataset, which we refer to as \textit{seed samples}, several samples are generated removing views.
Given a sample $s$ comprising data obtained from $n$ different views ($s={d_1, d_2, ..., d_n}$), $m$ new samples can be generated by selecting subsets of the views.
The subsets are chosen randomly from all possible combinations that can be obtained with the different number of views (from $1$ to $n$).
For example, suppose a seed sample $s$ contains data from $5$ views ($s=\{d_1, d_2, d_3, d_4, d_5\}$), and we want to generate 3 new samples from $s$, the following samples could be randomly selected from the possible view combinations and added to the dataset: $\{d_1, d_3, d_4\}$, $\{d_2, d_5\}$, $\{d_3\}$.
The new data generated by this process are used as the input of the two networks.
However, in the case of the pose estimation network, for each generated sample, the whole data of its seed sample is used in the computation of the loss (equation \ref{mlp_loss}), as losing self-supervision information would not be beneficial.

\section{Experimental results} \label{sec:results}
Our 3D multi-human pose estimation system has been tested using the CMU Panoptic Studio dataset~\cite{panoptic} and a dataset generated at Aston University's Autonomous Robotics and Perception Laboratory for the purpose of this work.
The experiments that are presented in this section provide empirical evidence that our approach performs faster than other state-of-the-art algorithms, with comparable accuracy, and most importantly, does not require annotated datasets.
Finally, we provide evidence that the proposed HPE can successfully work on an autonomous robot.
\par

To perform these experiments, the two neural models were trained for each dataset using a train/validation/test split to prevent data leakage.
For the matching model we use a Graph Attention Network (GAT)~\cite{Velickovic2017} with $4$ hidden layers.
The hidden layers are composed of $[40, 40, 40, 30]$ hidden units and $[10, 10, 8, 5]$ attention heads.
LeakyReLU and Sigmoid are used as activation functions of the hidden and output layers, respectively.
The MLP-based pose estimator has 7 hidden layers of $[3072, 3072, 2048, 2048, 1024, 1024, 1024]$ hidden units, using LeakyReLU for the activation of the hidden layers and linear activation in the output layer.
\par

\subsection{Datasets} \label{sec:datasets}
To test our proposal on the \textbf{CMU Panoptic dataset} and facilitate comparisons, we use the same sequences and views as VoxelPose~\cite{tu2020voxelpose}. 
Similarly, the four sequences used for testing VoxelPose were applied in our experiments.
The data with the 2D skeletons' information were obtained using the backbone model provided in the VoxelPose project. 
Using this model, the 2D coordinates of the humans' keypoints were detected from the images. 
Nevertheless, our training strategy requires that each sample of the data includes the information of only one person, so it was necessary to organise the detection results to provide individual human data.
To this end, the skeletons of the different views belonging to the same human were identified using the ground truth of the Panoptic sequences and grouped to obtain individual samples for each human.
\par

As mentioned in section~\ref{sec:method}, our proposal assumes a fixed configuration of cameras for both the training and inference phases.
All the cameras have to be calibrated according to a global frame of reference as a previous step of the training.
Nevertheless, this requirement is not strictly met in the Panoptic dataset, which entails some limitations in the comparison with the ground truth.
To illustrate the problem, table~\ref{tab:transaltion_datasets_panoptic} shows the translation vector to the global frame of reference of the five selected cameras for two of the datasets.
As can be observed, there are significant variations between the positions of the cameras in the two datasets, exceeding, in some cases, and for specific axes, 0.1m.
This implies that each sequence considers a different global frame of reference.
To overcome this limitation, we use the calibration file of one of the datasets (160224\_haggling1) for training and, for testing, the ground truth of each test dataset is transformed from the global reference frame of that dataset to the global reference frame used for training.
For applying that transformation, a specific camera is used as a common frame of reference for all the datasets.
Thus, the ground truth is first transformed from the global frame of reference of the dataset to the camera frame of reference and then from the camera frame of reference to the global frame of reference used for training.
Although transformation partly solves the problem of having different calibration data for each dataset, a residual error remains due to small variations of the intrinsic and inter-camera extrinsic parameters in the Panoptic sequences.
Even though this fact is detrimental for our approach in the comparison, the results are still comparable.

\begin{table*}[ht!]
\centering
\caption{Translation (in millimeters) between each camera and the global frame of reference for two sequences of the CMU Panoptic dataset.}
\label{tab:transaltion_datasets_panoptic}
% \resizebox{\textwidth}{!}{
\begin{tabular}{lcccccc}
\hline
             & \multicolumn{6}{c}{\textbf{Sequence}}             \\
\cmidrule(lr){2-7}             
            & \multicolumn{3}{c}{\textbf{160224\_haggling1}}                    & \multicolumn{3}{c}{\textbf{160422\_haggling1}}                   \\ 
\cmidrule(lr){2-4}\cmidrule(lr){5-7}            
            \multicolumn{1}{c|}{\textbf{Camera}}        & X & Y & \multicolumn{1}{c|}{Z}                    & X & Y & Z                   \\ \hline                         
\multicolumn{1}{l}{HD03} & \multicolumn{1}{|c|}{2087.71} & \multicolumn{1}{|c|}{-1510.89} & \multicolumn{1}{|c|}{1780.99} & \multicolumn{1}{|c|}{2015.19} & \multicolumn{1}{|c|}{-1512.49} & \multicolumn{1}{|c}{1789.8} \\ \hline
\multicolumn{1}{l}{HD06} & \multicolumn{1}{|c|}{-677.9} & \multicolumn{1}{|c|}{-3394.66} & \multicolumn{1}{|c|}{-1704.22} & \multicolumn{1}{|c|}{-641.81} & \multicolumn{1}{|c|}{-3398.11} & \multicolumn{1}{|c}{-1783.39} \\ \hline
\multicolumn{1}{l}{HD12} & \multicolumn{1}{|c|}{-76.23} & \multicolumn{1}{|c|}{-2392.45} & \multicolumn{1}{|c|}{2552.27} & \multicolumn{1}{|c|}{-173.19} & \multicolumn{1}{|c|}{-2395.36} & \multicolumn{1}{|c}{2494.95} \\ \hline
\multicolumn{1}{l}{HD13} & \multicolumn{1}{|c|}{-1840.95} & \multicolumn{1}{|c|}{-3393.29} & \multicolumn{1}{|c|}{143.76} & \multicolumn{1}{|c|}{-1860.28} & \multicolumn{1}{|c|}{-3393.9} & \multicolumn{1}{|c}{28.87} \\ \hline
\multicolumn{1}{l}{HD23} & \multicolumn{1}{|c|}{2343.16} & \multicolumn{1}{|c|}{-1526.22} & \multicolumn{1}{|c|}{-1433.97} & \multicolumn{1}{|c|}{2372.17} & \multicolumn{1}{|c|}{-1527.83} & \multicolumn{1}{|c}{-1417.28} \\ \hline
\end{tabular}
% }
\end{table*}

The\textbf{ ARP Laboratory dataset} was generated from 4 cameras attached to the walls of the laboratory and 2 additional cameras mounted on a robot. 
The robot was static and located at a fixed position during the generation of the dataset. 
All the cameras were calibrated in relation to a global frame of reference. 
A total of $18$ video sequences of single individuals moving were recorded. 
The sequences have variable lengths between $2'$ and $39'$. 
These sequences were used for training and testing separately the two models. 
Two additional sequences with groups of $2$ and $4$ people were recorded to test the whole system. 
These test sequences have a length of $3,43'$ and $2,58'$, respectively.

\subsection{Evaluation of the skeleton-matching module}
\label{subsec:evaluation_matching}

Since the goal of the skeleton-matching network is to group together the different views of a person, given an unknown number of people, it can be considered a clustering model.
Thus, the proposed matching technique can be evaluated through a set of clustering metrics.
Specifically, the following metrics have been used:

\begin{itemize}
    \item \textbf{Adjusted rand index} (ARI)~\cite{hubert1985}: estimates the similarity between two clusterings according to the number of pairs belonging to the same or different clusters. It is adjusted using a random model as baseline, ensuring a random clustering has a value close to 0. This score ranges between $-0.5$ (discordant clustering) and $1.0$ (perfect clustering).
    \item \textbf{Homogeneity} (H)~\cite{rosenberg-hirschberg-2007}: measures the homogeneity of the clusters. A cluster is considered homogeneous if it contains only members of the same class. It ranges between $0.0$ and $1.0$.
    \item \textbf{Completeness} (C)~\cite{rosenberg-hirschberg-2007}: measures the completeness of the clusters. A cluster is considered complete if all the members of the same class are assigned to the same cluster. It ranges between $0.0$ and $1.0$.
    \item \textbf{V measure} (Vm)~\cite{rosenberg-hirschberg-2007}: harmonic mean between homogeneity and completeness. This index quantifies the goodness of the clustering, considering both homogeneity and completeness. It ranges between $0.0$ and $1.0$.
\end{itemize}

These metrics have been applied to several skeleton matching networks trained for different numbers of views using the two datasets described in the previous section.
Table \ref{tab:matching_metrics_panoptic} shows the results for two, three, and five views using the four test sequences of the CMU Panoptic dataset.
For all the metrics, values close to $1$ are obtained regardless of the number of views.
% A slight improvement is observed for $3$ views, but overall the metrics show notable results for all the numbers of views.

\begin{table}[h!]
\centering
\caption{Metrics of the skeleton matching network for the CMU Panoptic dataset.}
\label{tab:matching_metrics_panoptic}
\begin{tabular}{|c||c|c|c|c|}
\hline
\textbf{No. of views}    & \textbf{ARI} & \textbf{H} & \textbf{C} & \textbf{Vm} \\ \hline \hline
\textbf{2}               & 0.9875 & 0.9968 & 0.9925 & 0.9943 \\ \hline
\textbf{3}               & 0.9977 & 0.9993 & 0.9981 & 0.9986 \\ \hline
\textbf{5}               & 0.9941 & 0.9978 & 0.9937 & 0.9956 \\ \hline
\end{tabular}%
\end{table}

The effectiveness of the proposed skeleton-matching network was also evaluated using the ARP Laboratory dataset. 
Two models, one with two views and the other with six views, were trained using ten of the eighteen sequences of single individuals.
The models were then tested on the remaining eight sequences, with a test dataset generated according to the multi-person dataset generation process detailed in section \ref{sec:skeleton_matching}.
This process provided the necessary ground truth to compute the evaluation metrics.

Table \ref{tab:matching_metrics_ARP} presents the results obtained from $2000$ samples in the generated dataset, which contained varying numbers of persons ranging from $1$ to $8$.
Similar to the CMU Panoptic dataset, the evaluation metrics demonstrated outstanding performance of the network for both the two and six views models.
Furthermore, it is noteworthy that for both datasets, the homogeneity values are nearly 1, indicating that the skeleton groups are predominantly comprised of views from the same individual.

\begin{table}[h!]
\centering
\caption{Metrics of the skeleton matching network for the ARP Laboratory dataset.}
\label{tab:matching_metrics_ARP}
\begin{tabular}{|c||c|c|c|c|}
\hline
\textbf{No. of views}    & \textbf{ARI} & \textbf{H} & \textbf{C} & \textbf{Vm} \\ \hline \hline
\textbf{2}               & 0.9770 & 0.9966 & 0.9886 & 0.9923 \\ \hline
\textbf{6}               & 0.9842 & 0.9974 & 0.9847 & 0.9905 \\ \hline
\end{tabular}%
\end{table}

\subsection{Evaluation of the multi-person 3D pose estimation system}
\label{subsec:evaluation_MLP}

The whole multi-person 3D pose estimation system has been evaluated for both, the CMU Panoptic and the ARP datasets.
For CMU Panoptic, the following metrics have been used:

\begin{itemize}
    % \item \textbf{Recall} (R): percentage of correctly estimated poses in relation to the total number of ground truth poses for a given distance threshold. Specifically, for a distance threshold $th$, an estimated pose is considered correct if the mean distance per keypoint to the ground truth is smaller than $th$.
    % \item \textbf{Precision} (P): percentage of correctly estimated poses in relation to the total number of detected poses for a given distance threshold.
    % \item \textbf{Average precision} (AP): average precision according to the precision-recall curve for a given distance threshold (as computed in \cite{tu2020voxelpose}).
    \item \textbf{Mean per joint position error} (MPJPE): mean distance (mm) per keypoint between detected and ground truth poses. 
    \item \textbf{Mean average precision} (mAP): mean of average precision over different distance thresholds (from $25$mm to $150$mm, taking steps of $25$mm). 
    \item \textbf{Mean recall} (mR): mean of recall over all the thresholds. 
    \item \textbf{Time for persons' proposals} ($t_{pp}$): mean time required for generating persons' proposals. In our approach, this time corresponds to the skeleton matching stage.
    \item \textbf{Time for 3D pose estimation} ($t_{3Dg})$: mean time required for estimating the 3D poses.
    \item \textbf{Time for 3D pose estimation per human} ($t_{3Di})$: mean time required for estimating the 3D pose of one person.
\end{itemize}
\par

To provide a comparison with other existing approaches, VoxelPose was trained using the same ten training Panoptic sequences. 
In addition, the results of our pose estimation model were compared with the 3D poses obtained by triangulation.
Specifically, for each pair of views of a person identified by the skeleton matching model, the 3D position of each visible keypoint was estimated by triangulating the 3D of its 2D coordinates.
If more than one estimation was obtained (\textit{i.e.}, the keypoint is visible from more than 2 cameras), the final 3D position for the keypoint was computed as the average of the individual estimations.

Regarding our proposal, we have used 3 different versions of the test dataset.
In the first version (\textit{D-detected}), the skeletons used as input contain the 2D coordinates of the keypoints detected from the backbone skeleton detector of VoxelPose. 
In the second version (\textit{D-projected}), the 2D positions of the detected keypoints have been replaced with the projected coordinates of the ground truth 3D keypoints.
Finally, in the third version (\textit{D-average}), the 2D positions of the keypoints have been computed as the average between the detected and projected 2D coordinates.
The main reason for using these 3 versions is the existing significant difference between the detected 2D keypoints and the ones obtained by projecting the ground truth 3D skeletons.
This fact can be observed in table \ref{tab:reprojection_error_gt}, where the reprojection error of the ground truth for the three datasets is shown.
To create this table, the ground truth 3D was projected into the images of the five cameras, considering the calibration data used to train our model, and compared with the 2D keypoints of the three datasets. 
As can be observed, there are significant differences among the reprojection errors considering the three datasets.
Thus, as expected, the largest reprojection error occurs on the \textit{D-detected} dataset, which indicates some divergences in the positions of the keypoints of the human body between the skeleton detector model and the ground truth of the Panoptic datasets.
In addition, certain differences are observed between the reprojected ground truth and the positions of the keypoints of the \textit{D-projected} dataset.
These differences are related to the different calibration data of each sequence of Panoptic.
Specifically, as mentioned in section \ref{sec:datasets}, the variations of the intrinsic and inter-camera extrinsic parameters of the sequences produce a remaining error that is reflected in the second row of table \ref{tab:reprojection_error_gt}.
In fact, compared with the \textit{D-avarage} dataset, cameras \textit{HD03} and \textit{HD23} present higher reprojection errors for the \textit{D-projected} dataset.

\begin{table}[ht!]
\centering
\caption{Reprojection error of the ground truth 3D for the three versions of the test dataset using CMU Panoptic.}
\label{tab:reprojection_error_gt}
% \resizebox{\columnwidth}{!}{
\begin{tabular}{lccccc}
\hline
             & \multicolumn{5}{c}{\textbf{Camera}}             \\
\cmidrule(lr){2-6}
\textbf{Dataset} & HD03 & HD06 & HD12 &  HD13 & HD23 \\ \hline \hline
D-detected & 7.01 & 10.73 & 7.63 & 10.71 & 6.37 \\ \hline
D-projected & 3.81 & 1.08 & 2.19 & 2.28 & 4.74 \\ \hline
D-average & 3.65 & 5.12 & 3.92 & 6.06 & 3.69 \\ \hline

\end{tabular}
% }
\end{table}

Table \ref{tab:mpjpe_and_time} shows a summary of the accuracy and time metrics obtained for VoxelPose, triangulation, and our proposed method across the four test sequences of the CMU Panoptic dataset.
The last three rows of the table correspond to our model's performance on the three dataset variations (\textit{D-detected}, \textit{D-projected}, and \textit{D-average}).

%%% PERCENTAGE OF COMPLETE POSES OF TRIANGULATION: 86.92
\begin{table*}[h!]
\centering
\caption{Accuracy and time metrics of VoxelPose, triangulation, and our proposal using the CMU Panoptic dataset.}
\label{tab:mpjpe_and_time}
% \resizebox{\columnwidth}{!}{
\begin{tabular}{lcccccc}
 \hline
\multicolumn{1}{c|}{\textbf{Method}}   & MPJPE & mAP & mR & $t_{pp}$ & $t_{3Dg}$ & $t_{3Di}$     \\ \hline  \hline                       
\multicolumn{1}{c|}{VoxelPose}   & 17.97 & 96,61 & 97,41 & 135.92 & 169.99 & 50.53     \\ \hline                                      
\multicolumn{1}{c|}{Triangulation}   & 22.63 & 76,99 & 85,10 & 32.56 &  10.06 & 2.99    \\ \hline                                      
\multicolumn{1}{c|}{Ours-detected}   & 26.06 & 89,25 & 92,63 & 31.67 & 19.65 & 5.83     \\ \hline                                      
\multicolumn{1}{c|}{Ours-projected}   & 17.84 & 96,23 & 97,76 & 31.96 & 19.94 & 5.89     \\ \hline                        
\multicolumn{1}{c|}{Ours-average}   & 19.77 & 95,67 & 97,39 & 32.22 & 19.81 & 5.85     \\ \hline                                      
\end{tabular}
% }
\end{table*}

\begin{figure*}[ht!]
\centering
  \begin{subfigure}{0.96\textwidth}
    \centering
    \frame{\includegraphics[width=0.75\textwidth]{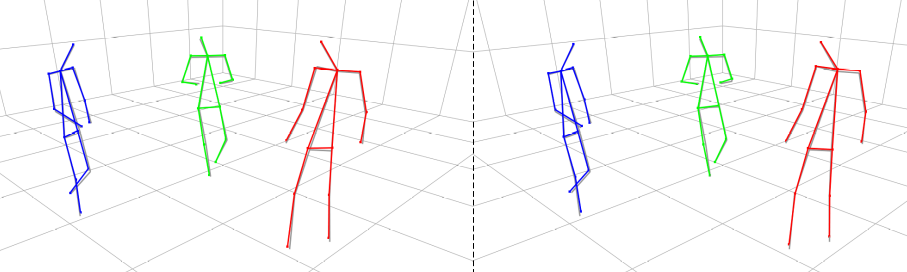}}
  \end{subfigure}
  \par\medskip
  \begin{subfigure}{0.96\textwidth}
    \centering
    \frame{\includegraphics[width=0.75\textwidth]{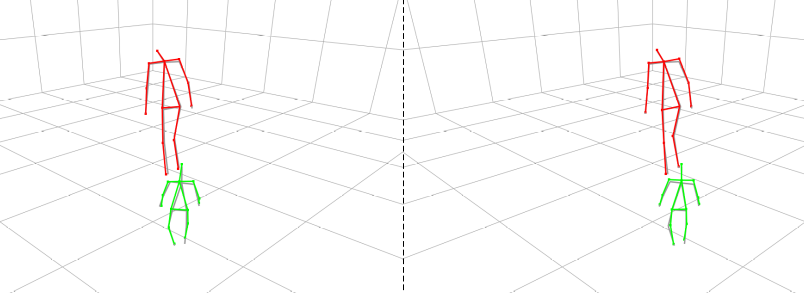}}
  \end{subfigure}
  \caption{Pose estimation results for 2 samples of the test sequences using our model (left images) and triangulation (right images). The ground truth is shown in gray. Triangulation provides complete poses in the 2 samples.}
  \label{fig:results_complete}
\end{figure*}

\begin{figure*}[h!]
\centering
  \begin{subfigure}{0.96\textwidth}
    \centering
    \frame{\includegraphics[width=0.75\textwidth]{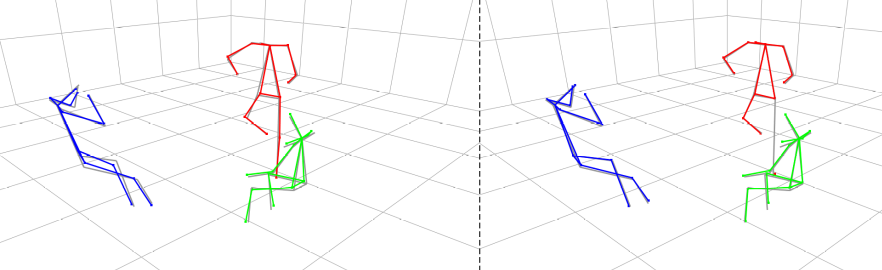}}
  \end{subfigure}
  \par\medskip
  \begin{subfigure}{0.96\textwidth}
    \centering
    \frame{\includegraphics[width=0.75\textwidth]{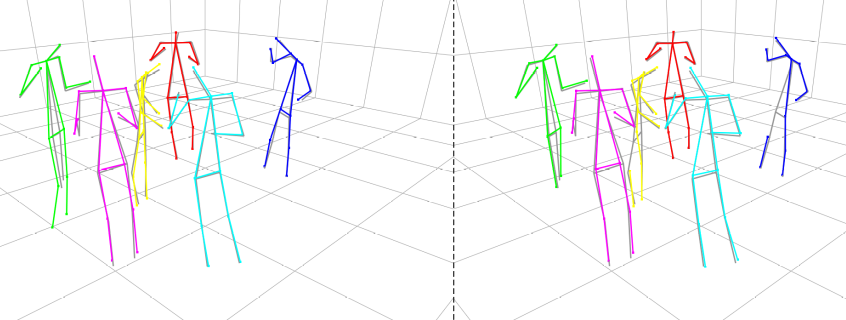}}
  \end{subfigure}
  \caption{Pose estimation results for 2 samples of the test sequences using our model (left image) and triangulation (right image). The ground truth is shown in gray. In these samples, triangulation cannot provide complete poses due to an insufficient number of views for some keypoints.}
  \label{fig:results_incomplete}
\end{figure*}

Regarding accuracy, VoxelPose and our model for the \textit{D-projected} dataset have similar performance, even though our model was trained with detected data.
Additionally, our model's results on the \textit{D-average} dataset for \textit{MPJPE}, \textit{mAP}, and \textit{mR} are quite comparable to those of VoxelPose.
The highest value of the \textit{MPJPE} is produced for our model with the \textit{D-detected} variation of the test dataset.
As mentioned earlier, this is due to the divergences between 2D detected and ground truth projected coordinates.
Nevertheless, triangulation yields a similar mean position error despite its computation only considering the keypoints for which triangulation can be applied, that is, the keypoints visible from two or more cameras.
Furthermore, triangulation performs the worst in terms of \textit{mAP} and \textit{mAR}.
This is due to the fact that triangulation does not always yield complete pose estimations. 
Figures \ref{fig:results_complete} and \ref{fig:results_incomplete} provide examples of complete and incomplete results using triangulation.
Figure \ref{fig:results_complete} showcases some samples where all the keypoints for every person in the scene can be estimated by triangulation.
In these cases, the estimated poses provided by our model (images on the left) and triangulation (images on the right) are very close to the ground truth poses (shown in gray).
However, in figure \ref{fig:results_incomplete}, some poses cannot be entirely determined by triangulation, as there are keypoints that are not visible from two or more cameras.
In such scenarios, our model provides complete estimates for all poses, with minimal differences from the ground truth.
Interestingly, the second scenario of figure \ref{fig:results_incomplete} presents a situation where the ground truth is incomplete (green skeleton). 
It can be seen how our model provides a realistic pose despite the lack of information.

In terms of computational time, VoxelPose takes an average of $305.91$ ms for the whole estimation process, which is almost $6$ times longer than the time required by our proposal.
The metrics of table \ref{tab:mpjpe_and_time} do not include the time required for skeleton detection, which may vary depending on the specific detector.
However, efficient solutions for detection do exist, such as \textit{trt-pose} which can perform at $251$ FPS on Jetson Xavier~\cite{trtpose}. 
Furthermore, skeleton detection for all the views can be run in parallel, making the timing roughly independent of the number of views.
Thus, assuming skeleton detection can be achieved at $30$ FPS, our proposal still runs more than $3$ times faster than VoxelPose.

Besides the aforementioned benefits regarding real-time execution, our proposal provides a more general solution to the problem than the existing alternatives.
The fact that \textbf{no ground truth is required} to train the two models makes our proposal easily replicable, regardless of the space, organization and extension.

\begin{figure*}[h]
\centering
  \begin{subfigure}{\textwidth}
    \centering
    \frame{\includegraphics[width=0.9\textwidth]{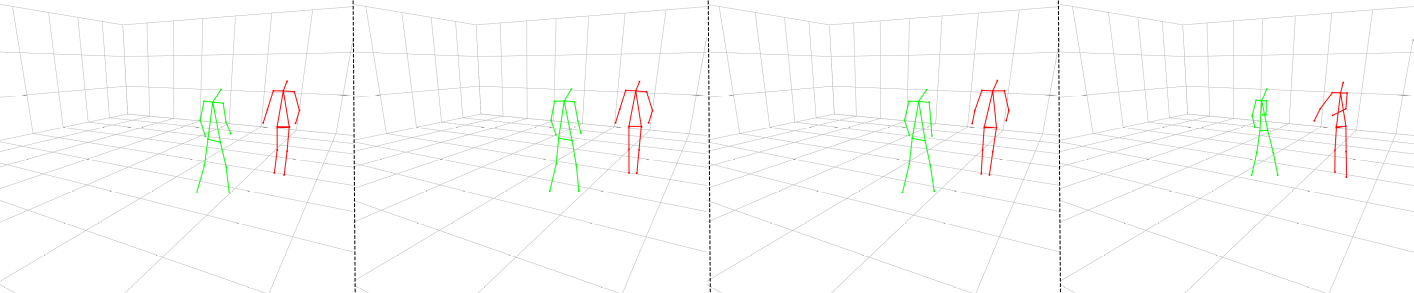}}
    % \caption{...}
  \end{subfigure}
  \par\medskip
  \begin{subfigure}{\textwidth}
    \centering
    \frame{\includegraphics[width=0.9\textwidth]{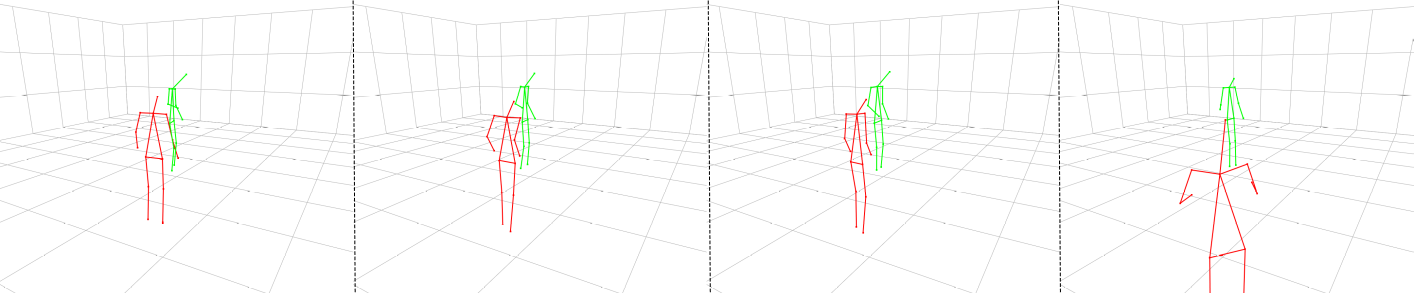}}
    % \caption{...}
  \end{subfigure}
  \par\medskip
  \begin{subfigure}{\textwidth}
    \centering
    \frame{\includegraphics[width=0.9\textwidth]{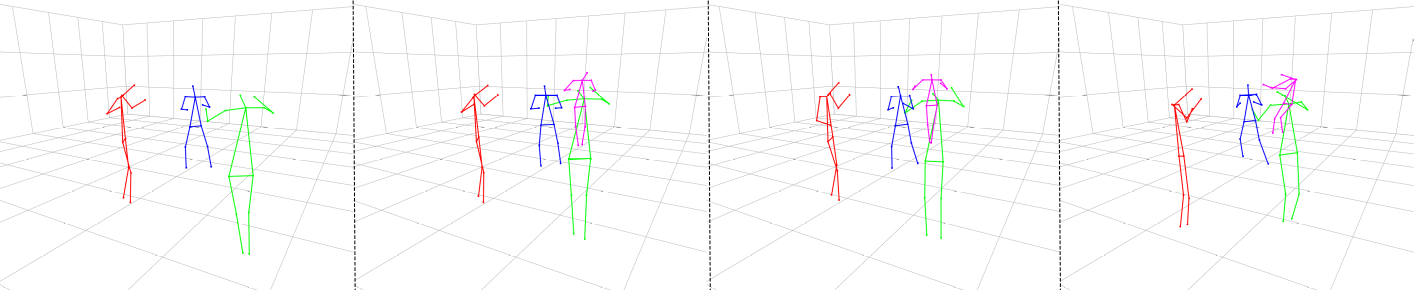}}
    % \caption{...}
  \end{subfigure}
  \par\medskip
  \begin{subfigure}{\textwidth}
    \centering
    \frame{\includegraphics[width=0.9\textwidth]{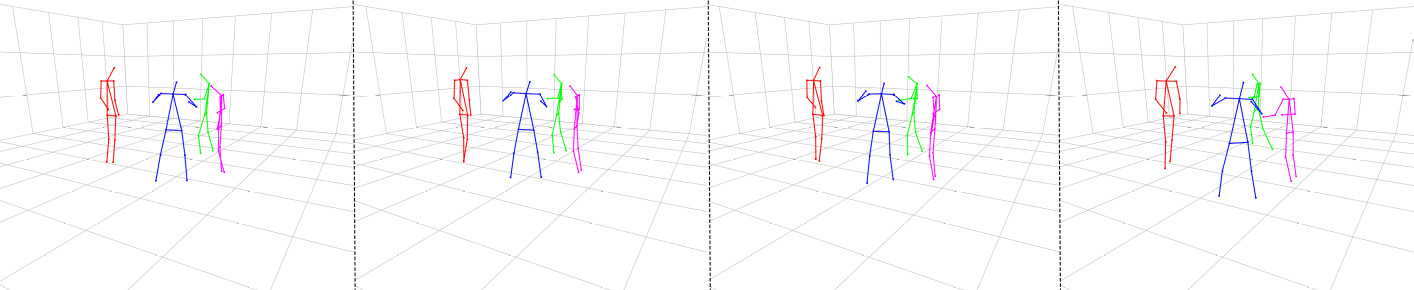}}
    % \caption{...}
  \end{subfigure}  
  \caption{Pose estimation results from our proposal of four samples of the ARP Laboratory multi-person sequences. From left to right, the results correspond to the models $M_{1/6}$, $M_{2/6}$, $M_{6/6}$, and $M_{2/2}$.}
  \label{fig:results_lab}
\end{figure*}

The proposed multi-person 3D pose estimation system has also been evaluated using the ARP Laboratory dataset.
Since this dataset does not include a ground truth, the accuracy metrics employed for the CMU Panoptic dataset can not be applied.
Instead, we use the reprojection error of the estimated 3D for all the cameras.
We trained four different models using different sets of cameras at training and inference times: $M_{6/6}$ that uses the six cameras for training and inference; $M_{2/6}$, which uses the six cameras for training and the two ones mounted on the robot for inference; $M_{1/6}$, that was trained with the six cameras but only uses one of the cameras of the robot at inference time; and $M_{2/2}$ that uses only the two cameras of the robot for training and inference.

Mean and median reprojection error for the four models and the six cameras (wall cameras: W0, W1, W2, and W3; robot cameras: R0 and R1) using the two ARP Laboratory sequences with $2$ and $4$ people are depicted in table \ref{tab:reprojection_error_LAB}.
The lowest reprojection error for the wall cameras is given by the model $M_{6/6}$.
This model is the most reliable one since it uses data from all the views.
Despite models $M_{2/6}$ and $M_{2/2}$ using the same cameras at inference time, there are very significant differences in their behavior, which is reflected in the reprojection errors of the cameras of the walls.
Especially, it can be observed a very high error of $M_{2/2}$ for the camera $W1$.
Such a large error is produced when there is limited visibility of a person from the cameras used by the model.
This is the case with the second sample of figure \ref{fig:results_lab}, where the model places the red skeleton far away from its actual position.
Besides this specific case, in general, the keyponts' positions estimated by model $M_{2/2}$ differ from the estimates of model $M_{6/6}$ as observed in that figure. 
In contrast, the model $M_{2/6}$ can estimate the pose of the person correctly, even though the input of both models is common.
Generally, the poses provided by the model $M_{2/6}$ are very similar to those provided by the model $M_{6/6}$, as demonstrated by both figure \ref{fig:results_lab} and the reprojection errors in table \ref{tab:reprojection_error_LAB}.
Finally, the model $M_{1/6}$ shows outstanding results considering it only receives information from one of the cameras of the robot (\textit{R0}).
The model is capable of predicting complete 3D poses with similar accuracy to model $M_{6/6}$, producing comparable reprojection errors to model $M_{2/6}$ \footnote{Bear in mind that, even though the reprojection errors are lower for $M_{1/6}$ than for $M_{2/6}$ in four of the six cameras, non-visible people from camera \textit{R0} (see the third sample of figure \ref{fig:results_lab}) are not considered in the error computation of model $M_{1/6}$.}. 

\begin{table*}[ht!]
\centering
\caption{Mean and median reprojection error in the 6 cameras of the ARP Laboratory for 4 models trained with different numbers of train and inference cameras.}
\label{tab:reprojection_error_LAB}
% \resizebox{\columnwidth}{!}{
\begin{tabular}{lcccc}
\hline
             & \multicolumn{4}{c}{\textbf{Model}}             \\
\cmidrule(lr){2-5}
\textbf{Camera} & $M_{1/6}$ & $M_{2/6}$ & $M_{6/6}$ &  $M_{2/2}$ \\ \hline \hline

W0 & 14.65 / 11.26 & 14.35 / 11.00 & 10.28 / 8.23 & 45.73 / 26.85 \\ \hline
W1 & 11.84 / 9.28 & 12.02 / 9.49 & 8.28 / 6.80  & 290.63 / 15.62 \\ \hline
W2 & 14.02 / 10.68 & 14.04 / 10.55 & 11.12 / 8.41  &  29.78 / 21.86 \\ \hline
W3 & 12.18 / 9.29 & 12.36 / 9.40 & 7.88 / 6.40 & 31.94 / 19.24 \\ \hline
R0 & 6.69 / 5.27 & 7.06 / 5.34 & 8.98 / 6.97 & 4.50 / 3.37 \\ \hline
R1 & 9.05 / 6.83 & 7.79 / 6.07 & 9.49 / 7.51 & 4.50 / 3.38 \\ \hline
\end{tabular}
% }
\end{table*}

The results of this experiment demonstrate that our system is capable of providing good estimations with a reduced number of cameras, by simply considering the information of an extended set of cameras during training.
This is a significant advantage for its application in autonomous robots, which is the focus of the next section.

\subsection{Evaluation in a mobile robot}
\label{subsec:evaluation_robot}
The aim of this experiment is to show the application of the proposed system in a mobile robot equipped with only two RGB cameras. 
The main goal is to endow the robot with the ability to estimate the complete 3D human poses with enough accuracy to enhance the interaction between them.
\par

Since the robot does not stay in a fixed location, the only visual information it can use is that provided by its two cameras.
In the case of a mobile robot, triangulation is less informative. The short baselines of robots' stereo systems rarely provide complete poses, and more importantly, they can cause small deviations in keypoints' image positions to produce large 3D errors.
Nevertheless, using only the data captured by the two cameras to train the pose estimation model does not provide reliable results, as shown in the previous section.
For this reason, we use the model $M_{2/6}$, which only requires the information of the two cameras on the robot at inference time, but uses the data from the four additional wall cameras during training.

\begin{figure*}[h!]
\centering
  \begin{subfigure}[t]{0.35\textwidth}
    \centering
    \includegraphics[width=1\textwidth]{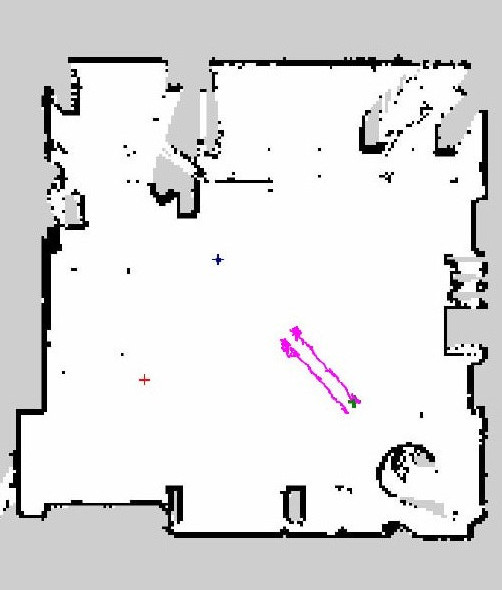}
    \caption{Trajectory of the ankles of the subject.}
    \label{fig:robot_static_map}
  \end{subfigure}
  \hspace{2mm}
  \begin{subfigure}[t]{0.6\textwidth}
    \centering
 \includegraphics[width=1\textwidth]{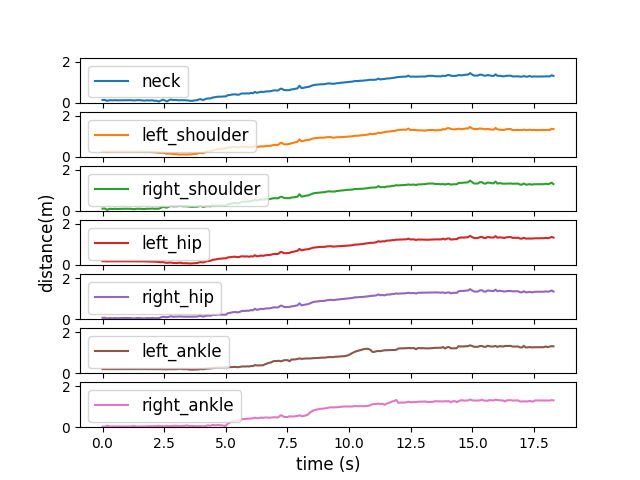}
    \caption{Distance from the initial position of the person to a subset of keypoints.}
    \label{fig:robot_static_graph}
  \end{subfigure}

  \caption{Experiment maintaining the robot in a fixed position while a person walks $1.5$ meters towards the robot. Only two RGB cameras are used at inference time.}
  \label{fig:robot_static_results}
\end{figure*}

We have conducted two different experiments to validate the effectiveness of the proposal. 
In the first experiment, the robot remains static facing the location of a person at $2.75$ meters from the front part of the robot.
The 3D pose of the person is recorded as they walk towards the robot in a straight line, covering a distance of approximately $1.5$ meters.
Figure \ref{fig:robot_static_map} illustrates the displacement of the person's two ankles (magenta lines), along with the position of the robot (dark blue cross).
The green cross in the map represents the initial position of the person and the red one the position of the global frame of reference. 
Figure \ref{fig:robot_static_graph} displays the displacement in meters of the coordinates (projected on the floor plane) of some representative keypoints from the initial position.
As can be seen, the traveled distance for all the keypoints goes from near $0$ to roughly $1.5$ meters.
In addition, the distance between symmetric keypoints presents only small variations along the entire route (e.g. the standard deviation is $2.3$cm for the hips and $0.64$cm for the eyes), which is indicative of the stability of the estimations.
\par

\begin{figure*}[h!]
\centering
  \begin{subfigure}[t]{0.35\textwidth}
    \centering
    \includegraphics[width=1\textwidth]{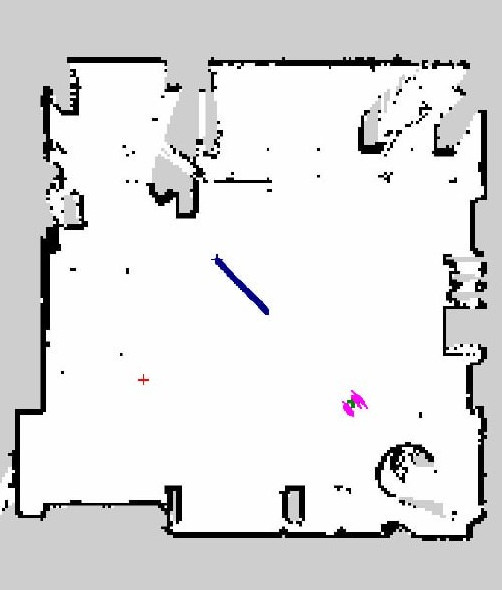}
    \caption{Trajectory of the robot.}
    \label{fig:person_static_map}
  \end{subfigure}
  \begin{subfigure}[t]{0.6\textwidth}
    \centering
 \includegraphics[width=1\textwidth]{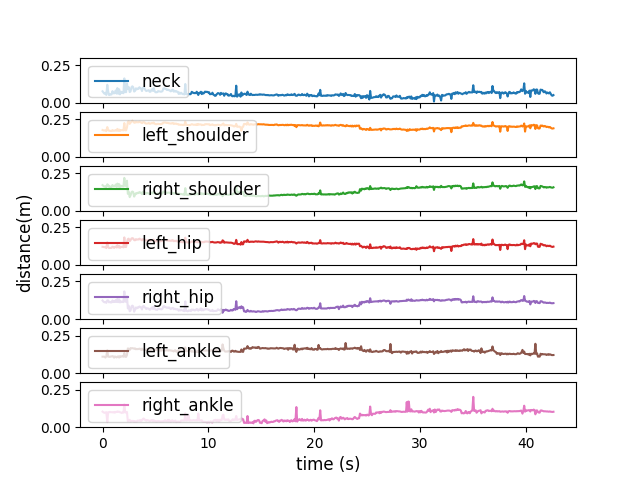}
    \caption{Distance from the initial position of the person to a subset of keypoints.}
    \label{fig:person_static_graph}
  \end{subfigure}
  \caption{Experiment where the person remains still while the robot moves $1$ meter towards the person. Only two RGB cameras are used at test time.}
  \label{fig:person_static_results}
\end{figure*}

The second experiment employs the same setup, with the person remaining stationary while the robot approaches them following a straight line covering a distance of $1$ meter.
Figure \ref{fig:person_static_map} shows the robot's path (thick blue line), extracted from its localisation system, and the position of the person's ankles, with the initial point situated between them.
Figure \ref{fig:person_static_graph} displays the graphs depicting the distances from the initial position of the person, at every instant, of the projection onto the floor of the same representative keypoints as in the previous experiment.
As shown in this figure, the distances remain almost constant, which is the expected result.
As in the previous experiment, the variations in the distances between symmetric keypoints are insignificant, with values from $1.8$cm for the ankles to $0.8$cm for the hips, which demonstrate the robustness and good accuracy of the model.

\section{Conclusions}
\label{sec:conclusions}
Multi-person 3D pose estimation is an important research field with multiple applications.
Deep learning is a powerful tool to learn human physiological priors.
Nevertheless, conventional deep learning solutions require large amounts of labelled data.
% To overcome this limitation, this work proposes a deep learning-based approach for the 3D multi-pose estimation problem that uses completely unannotated data.
We propose a GNN to identify the views of the different people in the scene and an MLP to estimate the complete 3D pose of each person.
Both networks are trained using completely unannotated data.
The unique requirement for the training of each network is that each element in the dataset corresponds to an individual person.
Both networks use information that can be directly obtained from the RGB images, so our approach only requires regular RGB cameras.
\par

Experimental results over our skeleton matching model for the CMU Panoptic and the ARP Laboratory datasets show outstanding model performance, which, as indicated in section~\ref{subsec:evaluation_matching}, yields nearly perfect values for all the clustering metrics.
\par
% Experimental results using the CMU Panoptic dataset show a mean per joint precision error of $26.06$mm, which can be considered more than acceptable for many applications.
% In addition, the $recall/precision$ ratio is very close to $1$, demonstrating the robustness of our skeleton matching model.
% This robustness is further demonstrated in Section \ref{subsec:evaluation_matching} where our model yields nearly perfect values for all the clustering metrics.

Regarding the accuracy of the 3D pose estimation model, in comparison to VoxelPose, our proposal shows slightly lower accuracy values over detected coordinates, with a mean per joint precision error of $26.06$mm.
Nevertheless, it is important to note, as investigated in Section \ref{subsec:evaluation_MLP}, that there is a significant difference between the detected 2D and the projection of the 3D ground truth in the CMU Panoptic dataset.
Using the projected 3D as a test set, without retraining the network, we obtain a mean per joint precision error of $17.84$mm (slightly better than VoxelPose error).
This disparity suggests that our 3D pose estimator's true precision might be higher than the one obtained in the initial comparison.
% This issue renders the evaluation somewhat unfair for our model and suggests that the real precision of our 3D pose estimator is higher. 
Furthermore, the computational complexity of our system is significantly lower than VoxelPose, making it an effective solution for real-time applications.
\par

The experiments conducted in Section \ref{subsec:evaluation_robot} illustrate the advantages of training our models with a subset of cameras for inference time. 
This approach enables the use of the system in a mobile robot with only two RGB cameras in real-time applications.
As demonstrated in these experiments, the system has sufficient precision to be used in social robotic applications.
\par

As future work, we aim to enhance the accuracy of our estimator model by refining the training with hyperparameter tuning. 
Additionally, we consider developing models trained with data including intrinsic and extrinsic camera parameters to pave the way to remove the need for scenario-specific training.
With this latter project, the model would be robot-agnostic, allowing users to simply mount the cameras on the robot, calibrate them, and run the system without any training.
\par


\bibliographystyle{IEEEtran}
\bibliography{bibliography.bib}

\end{document}